\documentclass[sigconf]{acmart}
\AtBeginDocument{%
  }

\setcopyright{acmlicensed}
\copyrightyear{2018}
\acmYear{2018}
\acmDOI{XXXXXXX.XXXXXXX}
\acmConference[Conference 'MM]{Make sure to enter the correct
  conference title from your rights confirmation email}{Nov. 10--14,
  2026}{Rio de Janeiro, Brazil}

\renewcommand\footnotetextcopyrightpermission[1]{}

\acmISBN{978-1-4503-XXXX-X/2026/04}

\acmSubmissionID{2427}

\usepackage{multirow}
\usepackage{booktabs}
\usepackage{colortbl}
\usepackage{threeparttable}
\usepackage{graphicx}
\usepackage{hyperref}

\usepackage{algorithm}

\usepackage{algpseudocode}

\usepackage[table]{xcolor}
\definecolor{ourgray}{RGB}{242,242,242}

\begin{document}

\title{Locality-Aware Density Control for Efficient Gaussian-based Image Representation}

\author{Jiacong Chen}
\orcid{0009-0008-3235-8281}
\affiliation{%
  \institution{Shenzhen University}
  \city{Shenzhen}
  \country{China}
}
\email{2210434045@email.szu.edu.cn}

\author{Qingyu Mao}
\affiliation{%
  \institution{Shenzhen University}
  \city{Shenzhen}
  \country{China}
}
\email{qingyu.mao@outlook.com}

\author{Xiandong Meng}
\affiliation{%
  \institution{Pengcheng Laboratory}
  \city{Shenzhen}
  \country{China}
}
\email{mengxd@pcl.ac.cn}

\author{Shuai Liu}
\affiliation{%
  \institution{Shenzhen University}
  \city{Shenzhen}
  \country{China}
}
\email{2205414001@email.szu.edu.cn}

\author{Chao Li}
\affiliation{%
  \institution{Harbin Institute of Technology (Shenzhen)}
  \city{Shenzhen}
  \country{China}
}
\email{lcc233@stu.hit.edu.cn}

\author{Fanyang Meng}
\affiliation{%
  \institution{Pengcheng Laboratory}
  \city{Shenzhen}
  \country{China}
}
\email{mengfy@pcl.ac.cn}

\author{Youneng Bao}
\authornote{Corresponding authors: Yongsheng Liang and Youneng Bao.}
\affiliation{%
  \institution{Shenzhen University}
  \city{Shenzhen}
  \country{China}
}
\email{baoyn@szu.edu.cn}

\author{Yongsheng Liang}
\authornotemark[1]
\affiliation{%
  \institution{Shenzhen University}
  \city{Shenzhen}
  \country{China}
}
\email{liangys@szu.edu.cn}

\begin{abstract}
2D Gaussian Splatting is an attractive direction for image representation due to its explicit formulation, fast rasterization, and favorable decoding efficiency.
The representation quality of this paradigm depends on the proper allocation of Gaussian capacity to the demanding regions.
However, existing methods fail to allocate Gaussian capacity efficiently during optimization: under-reconstructed content is often refined in a fragmented pixel-wise manner, while neighboring optimized Gaussians with similar attributes are redundantly retained.
This inefficiency motivates the need for a density control framework that jointly addresses insufficient allocation in under-reconstructed regions and redundant allocation in over-reconstructed regions.
Our key insight is that this framework should exploit two complementary forms of locality: the local continuity of reconstruction errors in image space for improved Gaussian allocation, and the local similarity of neighboring Gaussians in Gaussian space for redundant elimination. 
Based on this insight, we propose Locality-Aware Density Control (LocoADC), a plug-and-play framework that improves Gaussian capacity utilization through Region-wise Gaussian Densification (RGD) and Similarity-Driven Gaussian Merging (SDGM) strategies, together with a local color consistency constraint for more reliable merging.
Extensive experiments on diverse datasets show that LocoADC consistently improves multiple baselines by enabling more effective local Gaussian allocation, including a 2.93 dB PSNR gain over GI on the CLIC dataset under the same 30k Gaussian budget.
Code is available at: \textit{https://github.com/ChenJiaCong-1005/LocoADC}.
\end{abstract}


\keywords{Image Representation, Gaussian Splatting, Density Control}

\maketitle


\section{Introduction}
\label{sec:introduction}

\begin{figure}[t]
  \centering
  \includegraphics[width=0.48\textwidth]{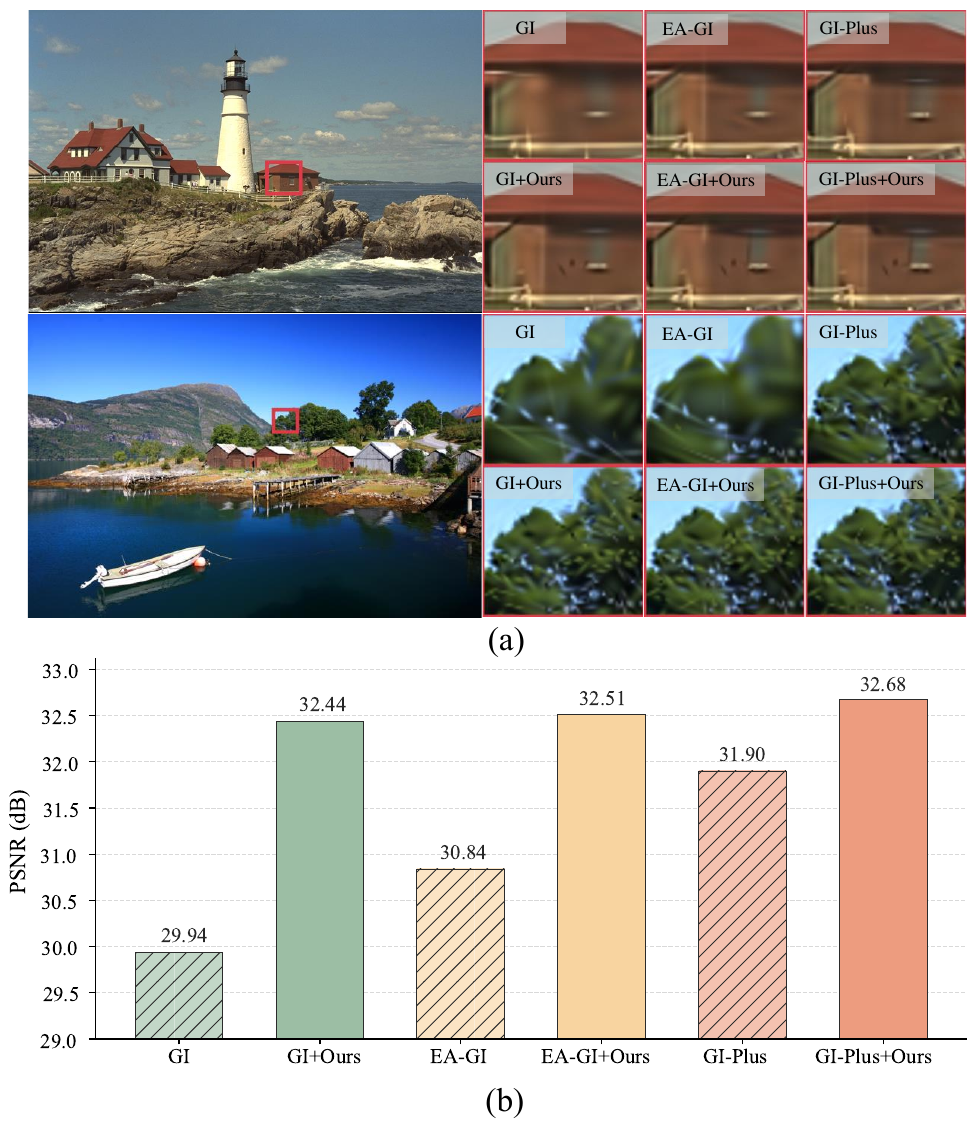}
  \caption{(a) Visual comparisons show that our method produces clearer and more faithful local details. (b) Quantitative results on the Kodak dataset~\cite{kodak} further demonstrate consistent PSNR improvements over the corresponding baselines.
}
  \label{Fig: Fig1}
\end{figure}

Image representation is a fundamental problem in computer vision and signal processing, with broad applications in image compression~\cite{dupont2021coin, strumpler2022implicit}, restoration~\cite{czerkawski2024neural, song2025triple}, and efficient visual content delivery.
A desirable image representation method should balance visual fidelity with optimization efficiency, compact storage, and fast decoding.
Recent years have witnessed rapid progress in implicit neural representations (INRs)~\cite{sitzmann2020implicit}, which model images as continuous functions and achieve impressive reconstruction quality. 
However, their strong performance typically relies on relatively large neural networks, leading to substantial training memory consumption and prolonged decoding time~\cite{zeng2025instant}.

To address these limitations, Gaussian-based image representation~\cite{zhang2024gaussianimage} has emerged as a promising direction by representing an image as a set of explicit 2D Gaussians and rendering it through efficient rasterization. 
This paradigm significantly reduces training memory consumption and achieves real-time decoding speed.
Building on this paradigm, subsequent works~\cite{zeng2025instant, chen2026entropy, liang2025structure} propose content-aware initialization strategies that leverage image priors to allocate more Gaussians to high-complexity regions before training. 
More recently, GI-Plus~\cite{li2026gaussianimage++} further introduced a distortion-driven densification during optimization to refine under-reconstructed pixels.

Despite these advances, current methods still lack an effective way to regulate Gaussian capacity during optimization.
This representation paradigm relies on multiple spatially neighboring and overlapping Gaussians to jointly model local image content that exhibits continuous structure. 
Therefore, representation efficiency depends on the proper allocation of local Gaussians to capture such content.
Existing methods lack effective locality-aware density control: initialization-based methods improve the starting allocation but cannot regulate evolving local budget imbalance during optimization, while existing densification methods mainly refine high-error responses in a pixel-wise manner but overlook local Gaussian redundancy.
As a result, representation quality depends on whether Gaussian capacity is properly allocated during optimization.



\begin{figure}[t]
  \centering
  \includegraphics[width=0.48\textwidth]{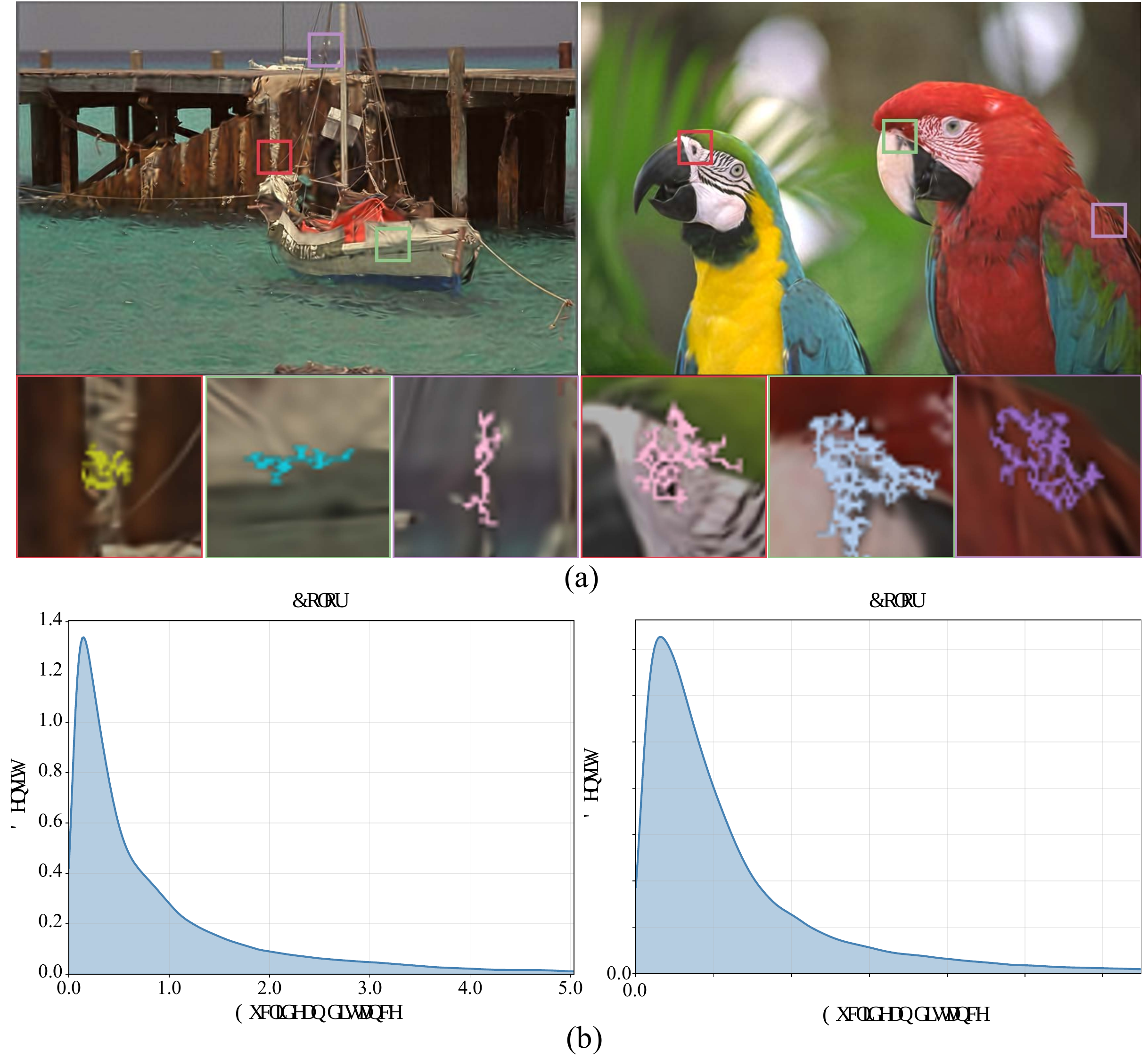}
  \caption{Locality of Gaussian-based image representation. (a) Reconstruction errors exhibit clear local continuity, where under-reconstructed pixels tend to form spatially coherent regions rather than isolated responses. (b) Spatially nearby Gaussians often show strong local color similarity, as evidenced by the concentration of small pairwise color differences. 
}
  \label{Fig: analysis}
\end{figure}

To better understand and exploit the locality, we analyze Gaussian-based image representation from two perspectives: image content and Gaussian space.
From the image-content perspective, challenging structures such as edges and textures usually exhibit strong local continuity, and the resulting reconstruction errors therefore form coherent local patterns rather than isolated pixel artifacts, as shown in Fig.~\ref{Fig: analysis}(a).
This observation suggests that densification should account for such locally continuous reconstruction errors in a region-aware manner, rather than treating high-error pixels independently.
From the Gaussian-space perspective, neighboring Gaussians often become highly similar in appearance and reconstruction contribution (Fig.~\ref{Fig: analysis}(b)), indicating that multiple Gaussians may repeatedly represent similar local content within the same region.
This local similarity indicates that these regions can be represented with fewer Gaussians by merging redundant neighboring Gaussians.


Based on this insight, we propose Locality-Aware Density Control (LocoADC), a plug-and-play framework for Gaussian-based image representation. 
Specifically, we introduce a Region-wise Gaussian Densification (RGD), which allocates Gaussians according to locally coherent distortion structures for more effective refinement of under-reconstructed regions. We further develop a Similarity-Driven Gaussian Merging (SDGM), which identifies and merges redundant neighboring Gaussians under appearance and stability constraints to improve representation efficiency. 
To support reliable merging, we additionally introduce a local color consistency constraint that stabilizes similarity estimation among neighboring Gaussians. 
As shown in Fig.~\ref{Fig: Fig1}, LocoADC effectively improves Gaussian capacity allocation during optimization by locally allocating Gaussians to under-reconstructed regions in region-wise manner and merging redundant Gaussians in over-reconstructed ones, without changing the underlying representation or decoding pipeline.

Extensive experiments on Kodak~\cite{kodak}, DIV2K $\times 2$~\cite{agustsson2017ntire}, and CLIC~\cite{toderici2020workshop} demonstrate that LocoADC consistently improves GI~\cite{zhang2024gaussianimage}, EA-GI~\cite{chen2026entropy}, and GI-Plus~\cite{li2026gaussianimage++}, while yielding more effective local Gaussian allocation. 
In particular, under the same 30k-Gaussian budget on the high-resolution CLIC dataset, LocoADC improves GI from 32.50 dB to 35.43 dB, achieving a 2.93 dB PSNR gain, and still boosts GI-Plus from 35.35 dB to 36.48 dB.

The main contributions of this paper are summarized as follows:
\begin{itemize}
\item We identify that the core inefficiency of Gaussian-based image representation lies in \textbf{locality-unaware Gaussian budget allocation} during optimization, which manifests as insufficient allocation in under-reconstructed regions and redundant allocation in over-reconstructed regions.
\item We propose \textbf{LocoADC}, a plug-and-play locality-aware density control framework that combines Region-wise Gaussian Densification and Similarity-Driven Gaussian Merging strategies, together with a local color consistency constraint for reliable merging.
\item We validate the proposed framework on diverse benchmarks and across multiple Gaussian-based baselines, showing that LocoADC consistently improves reconstruction quality under comparable Gaussian budgets and more effective local Gaussian allocation.
\end{itemize}


\begin{figure*}[t]
  \centering
  \includegraphics[width=1.0\textwidth]{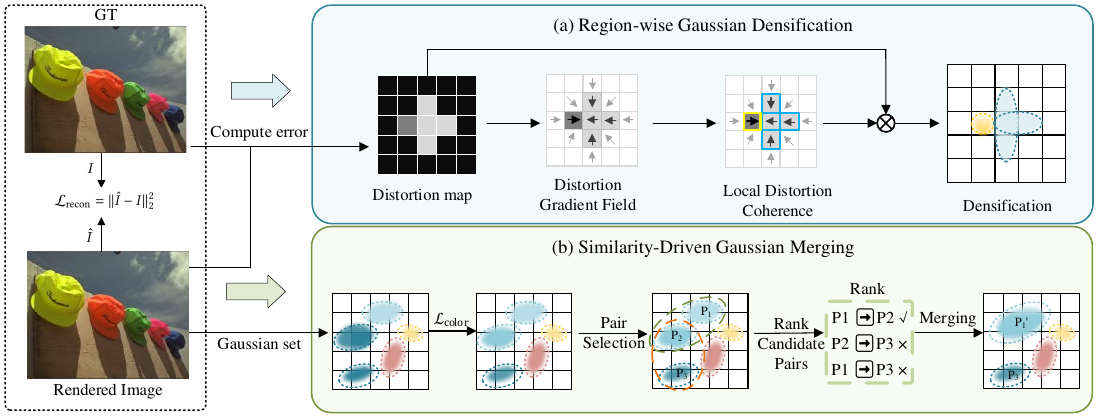}
  \caption{Illustration of the proposed LocoADC framework. Our LocoADC improves Gaussian allocation from two aspects: in under-reconstructed regions, it introduces Gaussians in a more proper manner through the region-wise Gaussian densification strategy, while in over-reconstructed regions, it merges locally similar Gaussians through the similarity-driven Gaussian merging strategy to avoid redundant representation.
}
  \label{Fig: main}
\end{figure*}


\section{Related Work}
\label{sec:related-work}
\subsection{Image Representation}
\label{sec:image representation}
Traditional image representations, such as JPEG~\cite{wallace1991jpeg} and JPEG2000~\cite{taubman2002jpeg2000}, represent images as discrete grid-based or transform-domain coefficients~\cite{cintra2011dct}.
While efficient and widely standardized, these approaches remain constrained by fixed transforms and hand-crafted modules, which limit their compression efficiency and visual quality.
Neural image representations~\cite{balle2018variational, cheng2020learned, he2022elic, BaoTJLLT25} employ autoencoder-based architectures within end-to-end optimization pipelines, achieving superior visual quality compared to traditional codecs.
However, these methods typically build upon fixed data organizations that provide little flexibility for adapting to image content.
Recently, Implicit Neural Representations (INRs)~\cite{sitzmann2020implicit, dupont2021coin} have emerged as a novel paradigm for image representation.
These methods model an image as a continuous mapping from spatial coordinates to RGB values, with the image content encoded in the parameters of a neural network.
Subsequent works further improve the representation capability of INRs by introducing positional encoding~\cite{tancik2020fourier} and designing activation functions~\cite{sitzmann2020implicit} that better capture high-frequency details and enhance visual quality.
However, the high performance of INRs often relies on large neural networks, and reconstructing each pixel requires a separate forward pass through the network, resulting in high GPU memory consumption and slow decoding speed.


\subsection{Gaussian-based Image Representation}
\label{sec:gaussian-image-representation}
3D Gaussian Splatting (GS)~\cite{kerbl20233d} represents 3D scenes by a collection of explicit 3D Gaussians with learnable attribute parameters.
As a pioneering work, GaussianImage (GI)~\cite{zhang2024gaussianimage} first extends GS into 2D image representation, which designs a compact 2D Gaussian with only 8 parameters and an accumulated blending-based rasterization without redundant computation.
However, this work fails to consider the imbalanced information distribution in the image, which employs a simple uniform initialization strategy to allocate Gaussians, resulting in under-reconstruction in complex regions.
To mitigate this limitation, InstantGI~\cite{zeng2025instant} proposes a generalizable network for efficient initialization and significantly reduces training time.
Fast 2DGS~\cite{wang2026fast} introduces a Deep Gaussian Prior to capture the spatial distribution of Gaussian primitives under different complexities. 
EA-GI~\cite{chen2026entropy} initializes Gaussians uniformly in the divided regions with approximately equal entropy, allowing each Gaussian to have equal information load.
StructureGI~\cite{liang2025structure} leverages spatial structural priors to design a structure-guided initialization.
Besides the initialization, MiraGe~\cite{waczynska2024mirage} utilizes adaptive density control in 3DGS to grow and prune the Gaussians, while GI-Plus~\cite{li2026gaussianimage++} introduces a distortion-driven densification to handle the under-reconstruction areas.
For ultra-high-resolution image representation, LIG~\cite{zhu2025large} fits these large images with a large number of Gaussians, and SmartSplat~\cite{li2026smartsplat} designs an adaptive and feature-aware framework to effectively support arbitrary image resolutions.
However, these methods largely optimize Gaussian representations in a pixel-wise manner and ignore the locality underlying both image structures and Gaussian distributions. 
As a result, locally continuous high-error regions may be fragmented during densification, while neighboring Gaussians with similar attributes may remain redundant after optimization. 
In contrast, our method explicitly models locality in both pixel space and Gaussian space, enabling region-wise Gaussian densification and redundancy-aware Gaussian merging within a unified framework.

\subsection{Density Control in 3D Gaussian Splatting}
\label{sec:density-control-in-3d-gaussian-splatting}
Existing density control methods~\cite{kerbl20233d, ye2024absgs, lee2024compact} in 3DGS for scene reconstruction include two representative directions: densification for under-reconstructed regions~\cite{ye2024absgs, yu2024gaussian, zhang2024pixel} and sparsification for redundant Gaussians~\cite{lee2024compact, hamdi2024ges, liu2024compgs, fan2024lightgaussian}.
Densification-based methods refine the under-reconstructed regions by iteratively splitting or cloning Gaussians based on positional gradient magnitudes.
Sparsification-based methods reduce the number of Gaussians based on various criteria (e.g., attribute-based~\cite{liu2024compgs, papantonakis2024reducing}, or optimization-driven~\cite{fan2024lightgaussian, wang2024end}) to avoid redundant representation while maintaining the visual quality.
These strategies are not suitable for Gaussian-based image representation, since low position gradients hinder reliable densification~\cite{li2026gaussianimage++} and directly removing Gaussians may degrade reconstruction quality (validated in Sec .~\ref {sec:ablation-study}).

\section{Methodology}
\label{sec:methodology}
We begin with the basics of Gaussian-based image representation, and then introduce our proposed Locality-Aware Density Control (LocoADC) framework, as illustrated in Fig.~\ref{Fig: main}.



\subsection{Preliminaries}
\label{sec:preliminaries}
GI~\cite{zhang2024gaussianimage} represents a 2D image as a set of 2D Gaussians $\mathbb{G}=\bigcup_{i=1}^N {g_i}$ by an accumulated summation rasterization.
Each Gaussian is characterized by a position vector $\mu \in \mathbb{R}^2$, a covariance matrix $\Sigma \in \mathbb{R}^{2 \times 2}$, and a color value $c \in \mathbb{R}^3$.
To ensure the positive semi-definiteness of $\Sigma$ during optimization, existing methods typically adopt three strategies: (1) Cholesky factorization~\cite{zhang2024gaussianimage}; (2) decomposing into a scaling matrix $S \in \mathbb{R}^{2 \times 2}$ and a rotation matrix $R \in \mathbb{R}^{2 \times 2}$~\cite{kerbl20233d}; (3) directly optimizing and pruning no-positive semi-definite cases~\cite{li2026gaussianimage++, zhu2025large}.

The learnable parameters of each Gaussian are optimized by differentiable rasterization.
To render an image, GI simplifies $\alpha$-blending with accumulated blending-based rasterization that eliminates the need to consider project transformation and depth information.
The color of $C(x)$ of a given pixel ${x}$ is rendered as
\begin{equation}
  C({x}) =  \sum_{i \in N} c_i \cdot \text{exp}(- \sigma_i),
\end{equation}
\begin{equation}
  \sigma_i = \frac{1}{2} d_i^T \Sigma^{-1} d_i,
\end{equation}
where $d_i \in \mathbb{R}^2$ is the distance between the corresponding pixel center and the Gaussian position.
Gaussian parameter optimization is supervised by the reconstruction loss $\mathcal{L}_\text{recon}$:
\begin{equation}
  \mathcal{L}_{\mathrm{recon}} = \| \hat{I}-I \|_2^2,
\end{equation}
where $\hat{I}$ and $I$ represent the rendered and ground truth images, respectively.

In addition to Gaussian parameter optimization, the Gaussian set can be dynamically updated during training through a density control framework \(\mathcal{C}\). This process, widely adopted in 3DGS~\cite{kerbl20233d}, can be generally formulated as
\begin{equation}
  \mathcal{C}(\mathbb{G}_t)=
\begin{cases}
\mathcal{D}(\mathbb{G}_t), & \Delta |\mathbb{G}_t| > 0, \\
\mathcal{S}(\mathbb{G}_t), & \Delta |\mathbb{G}_t| < 0 .
\end{cases}  
\end{equation}
Here, \(\mathcal{D}(\mathbb{G}_t)\) and \(\mathcal{S}(\mathbb{G}_t)\) denote the densification and sparsification operations applied to the Gaussian set \(\mathbb{G}_t\), respectively.
Density control provides a natural framework for efficient Gaussian allocation by densifying Gaussians in under-reconstructed regions \(\mathcal{D}(\mathbb{G}_t)\) and sparsifying them in over-reconstructed ones \(\mathcal{S}(\mathbb{G}_t)\).
However, most existing methods~\cite{chen2026entropy,zeng2025instant,liang2025structure} focus primarily on Gaussian initialization and overlook the allocation inefficiency that arises during optimization. 
GI-Plus~\cite{li2026gaussianimage++} takes a step further by densifying under-reconstructed pixels, but its refinement remains pixel-wise and one-sided. 
In contrast, our density control framework exploits locality to allocate Gaussians more efficiently and appropriately to the local regions, thereby enhancing the representation capability of Gaussian-based image representation.


\subsection{Region-wise Gaussian Densification}
\label{sec:structure-guided-regional-densification}
Guided by the above analysis, we develop the densification branch \(\mathcal{D}(\mathbb{G}_t)\) in our LocoADC framework as Region-wise Gaussian Densification (RGD) strategy (Fig.~\ref{Fig: main}(a)) to refine under-reconstructed regions. 
Unlike existing pixel-wise densification strategies that treat high-error pixels independently, our goal is to densify locally coherent distortion regions, enabling additional Gaussians to be allocated more effectively and efficiently to insufficiently represented content.

We first compute a distortion map to measure the reconstruction error of the rendered image, which can be defined as:
\begin{equation}
  \mathcal{E}(x)=\|I(x)-\hat{I}(x)\|_2,
\end{equation}
where $I(x)$ and $\hat{I}(x)$ represent the ground truth and rendered image at pixel $x$, respectively.
This distortion map provides a direct cue for locating high-error regions.
However, identifying under-reconstructed regions solely based on the distortion map remains insufficient, since it ignores the internal structural patterns of local reconstruction distortions.
To complement this information, we further extract the distortion gradient field $\mathbf{u}(x)$ from the distortion map using the Sobel operator:
\begin{equation}
  \mathbf{u}(x)=\frac{\left(\nabla_\text{x} \mathcal{E}(x),\nabla_\text{y} \mathcal{E}(x)\right)^\top}{|\nabla \mathcal{E}(x)|_2+\epsilon},
\end{equation}
where $\nabla_x \mathcal{E}(x)$ and $\nabla_y \mathcal{E}(x)$ denote the horizontal and vertical gradient of the distortion map $\mathcal{E}(x)$, respectively, and $\epsilon$ is a small constant introduced for numerical stability.
Based on the distortion gradient field, we further compute the local distortion coherence within each neighborhood to identify whether nearby reconstruction errors are spatially correlated. 
As the distortion gradient field captures the local variation of the distortion map, consistent gradient directions imply that neighboring high-error responses follow a shared spatial trend rather than appearing as isolated deviations. 
This local correlation indicates a coherent under-reconstructed region, which is more suitable for region-wise densification.
Accordingly, we define the local distortion coherence $D(x)$ at pixel $x$ as:
\begin{equation}
\label{eq:local coherence}
  D(x)=\left|\frac{1}{w^2}\sum_{q\in\mathcal{N}_w(x)} \mathbf{u}(q)\right|_2,
\end{equation}
where $\mathcal{N}_w(x)$ denotes a $w \times w$ window centered at pixel $x$.
Then, we jointly consider the distortion magnitude and the local distortion coherence to identify pixels that indicate under-reconstructed regions. 
Specifically, the candidate set is defined as
\begin{equation}
  \mathcal{C}_K=\operatorname{TopK}(\mathcal{E}(x)\cdot D(x)).
\end{equation}
Among the selected candidates, spatially neighboring pixels are first grouped into regional subsets, and one new Gaussian is allocated to each region and initialized at its center. 
If the densification budget remains available after regional refinement, the remaining budget is further allocated to isolated candidate pixels.

 \begin{table*}[htbp]
    \centering
    \normalsize
    \renewcommand{\arraystretch}{1.1}
    \caption{Quantitative comparison on the Kodak, DIV2K $\times 2$, and CLIC datasets. For fair comparison, the results of the compared methods are reproduced using their official codebases with 120,000 training iterations.}
    \label{Tab: comparison}
    \resizebox{1.0\textwidth}{!}{%
    \begin{tabular}{lcccc|cccc|cccc}
    \toprule
    \multirow{2}{*}{Methods} 
    & \multicolumn{4}{c|}{Kodak} 
    & \multicolumn{4}{c|}{DIV2K $\times 2$}
    & \multicolumn{4}{c}{CLIC} \\
    \cmidrule(lr){2-5} \cmidrule(lr){6-9} \cmidrule(lr){10-13}
    & PSNR(dB) & MS-SSIM & Params(M) & Training(s)
    & PSNR(dB) & MS-SSIM & Params(M) & Training(s)
    & PSNR(dB) & MS-SSIM & Params(M) & Training(s) \\
    \midrule
    
     \multicolumn{13}{l}{\textbf{INR-based}} \\
     SIREN~\cite{sitzmann2020implicit}
     & 26.50 & 0.8705 & 3.74 & 889
     & 30.01 & 0.9630 & 0.42 & 2242
     & 29.57 & 0.9386 & 0.57 & 4296\\
     SIREN+$T_{sym}$~\cite{zhang2025enhancing}
     & 32.80 & 0.9820 & 0.10 & 166 
     & 29.61 & 0.9712 & 0.10 & 261
     & 29.45 & 0.9392 & 0.10 & 824 \\
     EVOS~\cite{zhang2025evos}
     & 33.64 & 0.9809 & 0.10 & 106 
     & 30.38 & 0.9687 & 0.10 & 189
     & -- & -- & -- & -- \\
     \midrule
    
    \multicolumn{1}{l}{\textbf{GS-based}} 
    & \multicolumn{4}{c|}{Number of GS (5k)}
    & \multicolumn{4}{c|}{Number of GS (10k)}
    & \multicolumn{4}{c}{Number of GS (10k)}
    \\
    \cmidrule(lr){2-5} \cmidrule(lr){6-9} \cmidrule(lr){10-13}
    
    3D GS~\cite{kerbl20233d}
    & 27.22 & 0.942 & 0.3 & 436
    & 26.88 & 0.9499 & 0.59 & 722 
    & 27.29 & 0.9283 & 0.59 & 1980 \\
    MiraGe~\cite{waczynska2024mirage}
    & 29.83 & 0.9661 & 0.27 & 754
    & 29.89 & 0.9766 & 0.53 & 1117
    & 29.76 & 0.9507 & 0.53 & 3336 \\
    GI~\cite{zhang2024gaussianimage}
    & 29.94 & {0.9632} & 0.04 & 234.47
    & 29.71 & 0.9727 & 0.08 & 260.12
    & 29.35 & 0.9383 & 0.08 & 261.81 \\
    \rowcolor{yellow!12}
    GI+Ours 
    & 32.44 & 0.956 & 0.04 & 296.11 
    & 32.47 & 0.9713 & 0.08 & 335.01 
    & 31.52 & 0.9367 & 0.08 & 359.05 \\
    EA-GI~\cite{chen2026entropy}
    & 30.84 & 0.9614 & 0.04 & 247.8
    & 30.23 & 0.9653 & 0.08 & 301.31
    & 29.90 & 0.9274 & 0.08 & 305.23 \\
    \rowcolor{yellow!12}
    EA-GI+Ours
    & 32.51 & 0.9551 & 0.04 & 316.24 
    & 32.52 & 0.9711 & 0.08 & 341.53 
    & 31.57 & 0.9456 & 0.08 & 376.28 \\
    GI-Plus~\cite{li2026gaussianimage++}
    & 31.90 & 0.9614 & 0.04 & 247.75
    & 32.05 & {0.9742} & 0.08 & 283.28
    & 31.58 & 0.9430 & 0.08 & 284.86 \\
    \rowcolor{yellow!12}
    GI-Plus+Ours
    & {32.68} & 0.9596 & 0.04 & 313.88 
    & {32.84} & {0.9742} & 0.08 & 337.93 
    & {32.38} & {0.9468} & 0.08 & 386.27 \\
    
    \midrule
    
    \multicolumn{1}{l}{} 
    & \multicolumn{4}{c|}{Number of GS (10k)}
    & \multicolumn{4}{c|}{Number of GS (30k)}
    & \multicolumn{4}{c}{Number of GS (30k)}\\
    \cmidrule(lr){2-5} \cmidrule(lr){6-9} \cmidrule(lr){10-13}
    
    3D GS~\cite{kerbl20233d}
    & 29.96 & 0.968 & 0.59 & 457 
    & 31.29 & 0.9779 & 1.77 & 759 
    & 30.67 & 0.9634 & 1.77 & 2122 \\
    MiraGe~\cite{waczynska2024mirage}
    & 32.57 & 0.9839 & 0.53 & 749 
    & 35.29 & 0.9943 & 1.59 & 1141 
    & 33.18 & 0.9789 & 1.59 & 3367 \\
    GI~\cite{zhang2024gaussianimage}
    & 32.58 & 0.9824 & 0.08 & 258.11
    & 34.64 & 0.9926 & 0.24 & 287.15
    & 32.5 & 0.9727 & 0.24 & 282.77 \\
    \rowcolor{yellow!12}
    GI+Ours 
    & 35.60 & 0.9794 & 0.08 & 336.56 
    & 37.66 & 0.992 & 0.24 & 355.34 
    & 35.43 & 0.9731 & 0.24 & 385.95 \\
    EA-GI~\cite{chen2026entropy}
    & 33.87 & 0.9806 & 0.08 & 267.21 
    & 35.59 & 0.9896 & 0.24 & 346.78
    & 33.29 & 0.9651 & 0.24 & 365.38 \\
    \rowcolor{yellow!12}
    EA-GI+Ours
    & 35.68 & 0.9784 & 0.08 & 343.9 
    & 37.89 & 0.9913 & 0.24 & 359.74 
    & 35.54 & 0.9736 & 0.24 & 382.47 \\
    GI-Plus~\cite{li2026gaussianimage++}
    & 35.41 & {0.9832} & 0.08 & 283.2
    & 38.30 & 0.9940 & 0.24 & 290.85
    & 35.35 & 0.9736 & 0.24 & 292.37 \\
    \rowcolor{yellow!12}
    GI-Plus+Ours
    & {36.15} & 0.9826 & 0.08 & 354.8 
    & {39.09} & {0.9942} & 0.24 & 369.61 
    & {36.48} & {0.979} & 0.24 & 420.65 \\
    \bottomrule
    \end{tabular}%
    }
    
  \end{table*}

\subsection{Similarity-Driven Gaussian Merging}
\label{sec:similarity-driven-gaussian-consolidation}
As shown in Fig.~\ref{Fig: main}(b), we design the sparsification branch \(\mathcal{S}(\mathbb{G}_t)\) in LocoADC as Similarity-Driven Gaussian Merging (SDGM) strategy, which leverages local similarity in Gaussian space to merge redundant neighboring Gaussians into a more compact and efficient representation, rather than directly pruning them~\cite{liu2024compgs, fan2024lightgaussian}.
Specifically, to support reliable merging, we first impose a local color consistency constraint during training. 
Based on this, SDGM strategy identifies similar Gaussian pairs under appearance and overlap constraints, and then merges them to improve representation compactness while preserving image fidelity.

\textbf{Local Color Consistency Regularization.}
Before Gaussian merging, we impose a local color consistency constraint to encourage neighboring redundant Gaussians to be more consistent in appearance and thus easier to merge reliably.
Specifically, we formulate this regularization over the color differences between spatially nearby Gaussian pairs within each local neighborhood:
\begin{equation}
  \mathcal{L}_{\mathrm{color}} = \frac{1}{|\mathcal{P}|} \sum_{(i,j) \in \mathcal{P}} |\mathbf{c}_i - \mathbf{c}_j|_2^2,
\end{equation}
where $\mathcal{P}$ denotes the set of Gaussian pairs that are spatial neighbors within a local radius.
Since this regularization only requires spatially nearby Gaussian pairs for local consistency enforcement, we construct these pairs by $KNN$ search with a radius constraint in 2D space.
During the regularized training stage, the total loss is defined as:
\begin{equation}
  \mathcal{L}_{\mathrm{total}} = \mathcal{L}_{\mathrm{rec}} + \lambda_{\mathrm{color}} \mathcal{L}_{\mathrm{color}},
\end{equation}
where $\lambda_{\mathrm{color}}$ is typically set to 0.01.

\begin{figure*}[t]
  \centering
  \includegraphics[width=0.94\textwidth]{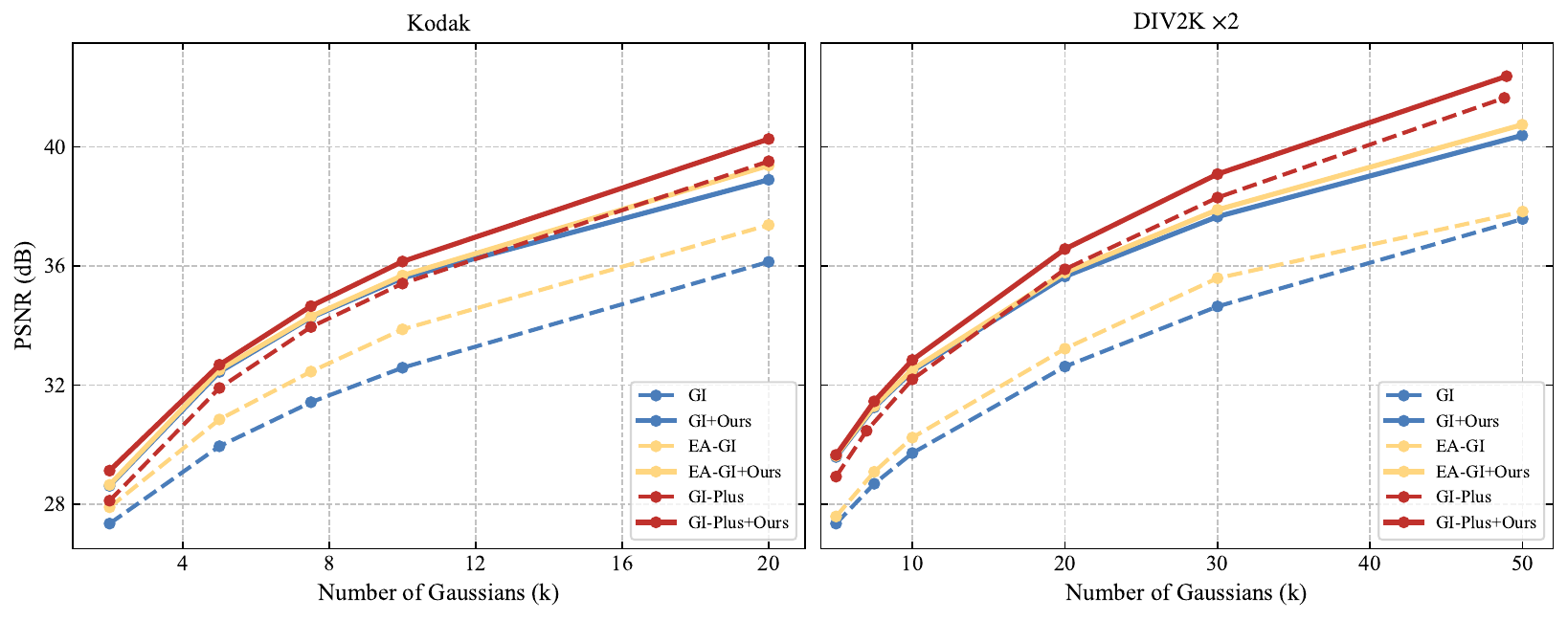}
  \caption{Comparison under different numbers of Gaussians on the Kodak and DIV2K $\times 2$ datasets.
  }
  \label{Fig: scalability_psnr}
\end{figure*}

\textbf{Similar Gaussian Pair Selection.}
To identify similar and mergeable Gaussians, we select candidate pairs within local neighborhoods under a Gaussian overlap constraint and a color similarity constraint.
We first filter out unstable Gaussians according to their accumulated optimization gradients, since Gaussians with large gradients often correspond to regions that are still under active correction and may incur large errors after merging:
\begin{equation}
  \frac{g_i}{\max_k g_k} < 0.5,
\end{equation}
where $g_i$ denotes the accumulated optimization gradient magnitude of Gaussian $i$.
Then, we identify spatially adjacent Gaussian pairs among the retained Gaussians based on Mahalanobis overlap:
\begin{equation}
  d_{\mathrm{M}}(\boldsymbol{\mu}_i, \boldsymbol{\mu}_j; \boldsymbol{\Sigma}_i)
= \sqrt{(\boldsymbol{\mu}_i-\boldsymbol{\mu}_j)^\top \boldsymbol{\Sigma}_i^{-1}(\boldsymbol{\mu}_i-\boldsymbol{\mu}_j)} < \tau_{\mathrm{M}},
\end{equation}
where $d_{\mathrm{M}}$ denotes the Mahalanobis distance, and $\tau_{\mathrm{M}}$ is the adjacency threshold.
This criterion treats two Gaussians as neighboring when one Gaussian center falls within the effective region of the other.
Then, we retain only Gaussian pairs with similar colors:
\begin{equation}
  |\mathbf{c}_i-\mathbf{c}_j|_2 < \tau_{\mathrm{color}},
\end{equation}
where $\tau_{\mathrm{color}}$ is the color similarity threshold.
The Gaussian pairs satisfying both criteria form a candidate set for the subsequent merging.
Since one Gaussian may appear in multiple candidate pairs, we further define a Pair Discrepancy Score to rank them:
\begin{equation}
  \mathcal{D}(i,j)=|\boldsymbol{\mu}_i-\boldsymbol{\mu}_j|_2^2+|\mathbf{c}_i-\mathbf{c}_j|_2^2+|\boldsymbol{\ell}_i-\boldsymbol{\ell}_j|_2^2,
\end{equation}
where $\boldsymbol{\ell}_i$ denotes the vectorized lower-triangular entries of $\Sigma_i$.
A lower discrepancy indicates a more compatible pair for merging. 
If a Gaussian is involved in multiple candidate pairs, we retain only the pair with the lowest discrepancy.

\textbf{Gaussian Merging.}
Given the selected Gaussian pairs, we perform Gaussian merging to improve representation efficiency. 
To reduce the degradation caused by merging, the merged Gaussian is updated to preserve the local representative content of the original pair.
Specifically, we update the merged Gaussian by area-weighted averaging of its position and color:
\begin{equation}
\label{eq:merging weight}
\begin{gathered}
  w_i = \frac{S_i}{S_i + S_j}, \quad w_j = \frac{S_j}{S_i + S_j}, \\
  \boldsymbol{\mu}_{\mathrm{merged}} = w_i \boldsymbol{\mu}_i + w_j \boldsymbol{\mu}_j, \\
  \mathbf{c}_{\mathrm{merged}} = w_i \mathbf{c}_i + w_j \mathbf{c}_j,
\end{gathered}
\end{equation}
where $S_i$ and $S_j$ denote the effective support areas of the two Gaussians, $w_i$ and $w_j$ are the corresponding merge weights.
To preserve the covered area of both original Gaussians, we initialize the covariance via second-moment matching:
\begin{equation}
\begin{gathered}
  \mathbf{M}_{\mathrm{merged}} = w_i \left(\boldsymbol{\Sigma}_i + \boldsymbol{\mu}_i \boldsymbol{\mu}_i^\top\right) + w_j \left(\boldsymbol{\Sigma}_j + \boldsymbol{\mu}_j \boldsymbol{\mu}_j^\top\right), \\
  \boldsymbol{\Sigma}_{\mathrm{merged}} = \mathbf{M}_{\mathrm{merged}} - \boldsymbol{\mu}_{\mathrm{merged}} \boldsymbol{\mu}_{\mathrm{merged}}^\top,
\end{gathered}
\end{equation}
where $\mathbf{M}_{\mathrm{merged}}$ denotes the area-weighted second moment of the original pair. 
This initialization allows the merged Gaussian to better cover the original area, reducing the merging error.
To avoid overly loose merging, we further constrain the effective support area of the merged Gaussian as:
\begin{equation}
  S_{\mathrm{merged}}  \leq \eta \cdot (S_i + S_j),
\end{equation}
where $S_{\mathrm{merged}}$ denotes the effective support area of the merged Gaussian and $\eta=1.5$ is the merging constraint ratio.

\section{Experiments}
\label{sec:experiments}

\subsection{Experimental Setup}
\label{sec:experimental-setup}
\textbf{Datasets and Evaluation Metrics.}
We evaluate our method on the image representation task using three widely used benchmarks: Kodak~\cite{kodak}, DIV2K~\cite{agustsson2017ntire}, and CLIC~\cite{toderici2020workshop}.
Kodak contains 24 images at a resolution of 768$\times$512. 
For DIV2K, we use the validation set and generate the test images by $2\times$ bicubic upsampling, resulting in resolutions ranging from 408$\times$1020 to 1020$\times$1020.
To further assess performance on high-resolution images, we also evaluate on the CLIC dataset, which contains 101 high-resolution images.
We employ PSNR and MS-SSIM~\cite{wang2003multiscale} to measure the representation quality.
We also report the parameter size and training time.

\textbf{Implementation Details.}
We select three typical Gaussian-based methods ($i.e.$, GI~\cite{zhang2024gaussianimage}, EA-GI~\cite{chen2026entropy}, and GI-Plus~\cite{li2026gaussianimage++}) as our baseline methods. 
Then, we enhance their performance by integrating our LocoADC framework. 
We optimize the Gaussian parameters for 120{,}000 iterations in total. Specifically, the first 50{,}000 iterations are used for normal fitting and densification, the next 20{,}000 iterations focus only on merging redundant Gaussians, and the final 50{,}000 iterations are used for fine-tuning with continued densification. 
We predefine a maximum number of Gaussians, initialize with half of this amount, and restrict densification to this limit.
Both densification and merging are performed every 5{,}000 iterations.
For a fair comparison, the results of all other methods are obtained using the same total number of iterations.
All experiments are conducted using NVIDIA RTX 4090 GPUs and PyTorch.
More implementation details are provided in \textbf{Appendix}.

 \begin{figure*}[t]
  \centering
  \includegraphics[width=.95\textwidth]{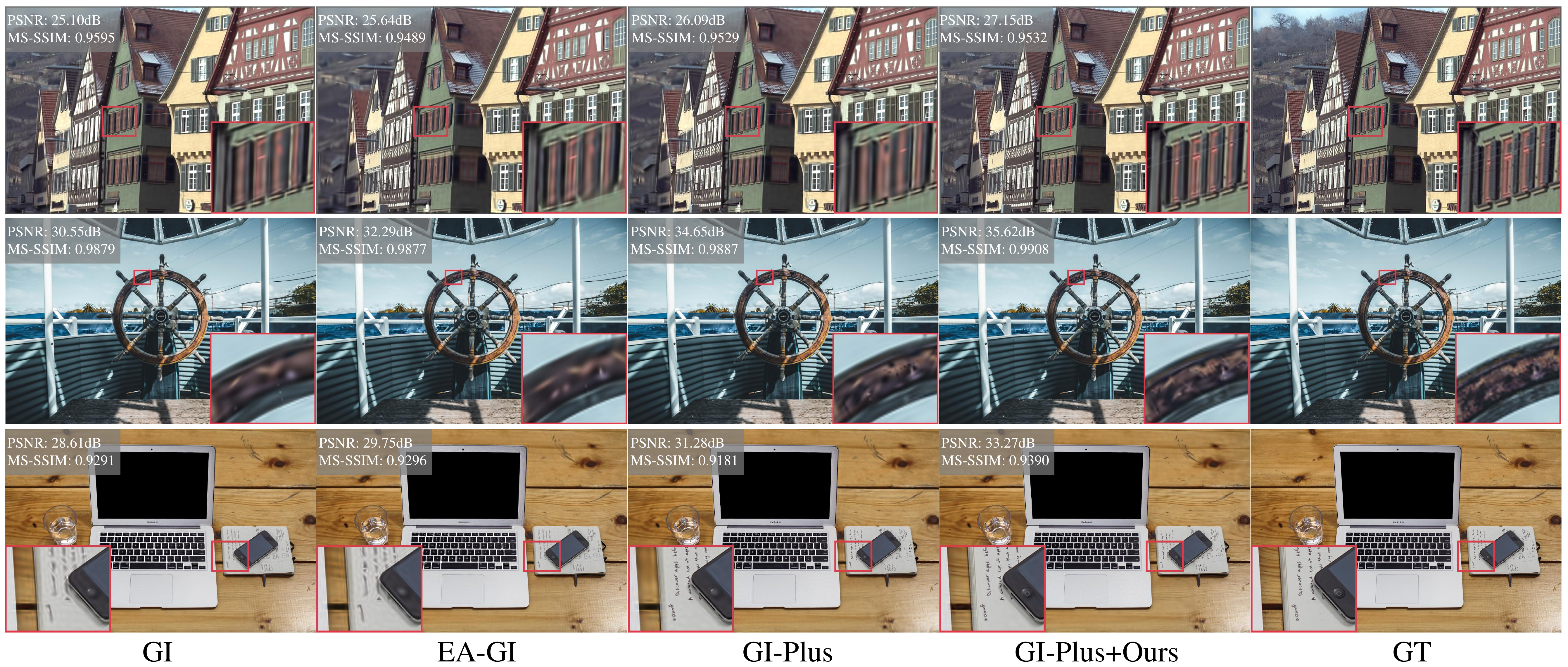}
  \caption{ Qualitative comparison of our approach with GI, EA-GI, and GI-Plus on the Kodak, DIV2K $\times 2$, and CLIC datasets. Our method consistently achieves improved visual quality across three datasets.
}
  \label{Fig: main comparison}
\end{figure*}

\subsection{Quantitative Results}
\label{sec:quantitative-results}
In addition to GI, EA-GI, and GI-Plus, we further provide quantitative comparisons with representative methods from INR-based and GS-based paradigms, including SIREN~\cite{sitzmann2020implicit}, SYM~\cite{zhang2025enhancing}, EVOS~\cite{zhang2025evos}, 3DGS~\cite{kerbl20233d}, and MiraGe~\cite{waczynska2024mirage}, to demonstrate the effectiveness of our method.
Tab.~\ref{Tab: comparison} presents the quantitative evaluation results on Kodak, DIV2K $\times 2$ and CLIC datasets under identical settings.
It is evident that our enhanced methods achieve consistently higher PSNR across all datasets and Gaussian budgets, while remaining competitive in MS-SSIM.
This improvement can be attributed to the efficient allocation of Gaussians in our framework, which densifies under-reconstructed regions in region-wise manner while reducing redundancy in over-reconstructed regions, thereby enhancing representation efficiency.
In addition, as shown in Fig.~\ref{Fig: scalability_psnr}, our method consistently achieves better results than the baseline across different numbers of Gaussians on both the Kodak and DIV2K $\times 2$ datasets.
It is worth noting that our method incurs additional training time, especially on the CLIC dataset. 
This overhead stems from distortion-map computation during densification, which is resolution-dependent, as well as from the merging process and local color consistency constraint, both of which involve neighboring-Gaussian search and thus scale with the number of Gaussians. 
Despite the additional training cost, the achieved performance improvements are substantial rather than simply resulting from longer optimization.
In the \textbf{Appendix}, more quantitative results are provided for a more detailed comparison.

\subsection{Qualitative Results}
\label{sec:qualitative-results}
 We visualize the reconstructed images to show the improvements achieved by our method.
 As shown in Fig.~\ref{Fig: Fig1}(a), compared with GI, EA-GI, and GI-Plus, the versions integrated with our method achieve substantially improved detail representation quality, especially for fine structures such as the house and leaves.
 In addition to Kodak and DIV2K $\times 2$, Fig.~\ref{Fig: main comparison} further includes comparisons on the high-resolution CLIC dataset.
 It can be observed that our method reconstructs clearer and more vivid structural textures. 
 This improvement can be attributed to the densification strategy guided by local structural distortions, enabling higher-quality recovery of structural information.
 Although some Gaussians are merged in our method, they are highly similar and can be adequately represented by fewer Gaussians.
 Therefore, our method still achieves competitive representation quality in smooth regions.
 
\begin{table}[t]
  \centering
  \small
  \setlength{\tabcolsep}{3.5pt}
  \renewcommand{\arraystretch}{1.2}
  \caption{Ablation study on evaluating RGD (Sec.~\ref{sec:structure-guided-regional-densification}) and SDGM (Sec.~\ref{sec:similarity-driven-gaussian-consolidation}) strategies on the Kodak, DIV2K $\times 2$, and CLIC datasets. The GI-Plus is selected as the baseline.}
  \begin{tabular}{cccc|ccc}
  \toprule
  GI-Plus & RGD & Merging & Color Cons. & Kodak & DIV2K $\times 2$ & CLIC \\
  \midrule
  $\checkmark$ &  &  &  & 31.9 & 32.05 & 31.58 \\
  $\checkmark$ & $\checkmark$ &  &  & 32.24  & 32.41 & 32.00 \\
  $\checkmark$ &  & $\checkmark$ &  & 32.34 & 32.51 & 32.01 \\
  $\checkmark$ &  & $\checkmark$ & $\checkmark$ & 32.52 & 32.69 & 32.19 \\
  \midrule
  $\checkmark$ & $\checkmark$ & $\checkmark$ & $\checkmark$ & \textbf{32.68} & \textbf{32.84} & \textbf{32.38} \\
  \bottomrule
  \end{tabular}
  
  \label{tab:ablation-psnr}
\end{table}

  \begin{figure*}[t]
      \centering
      \includegraphics[width=.95\textwidth]{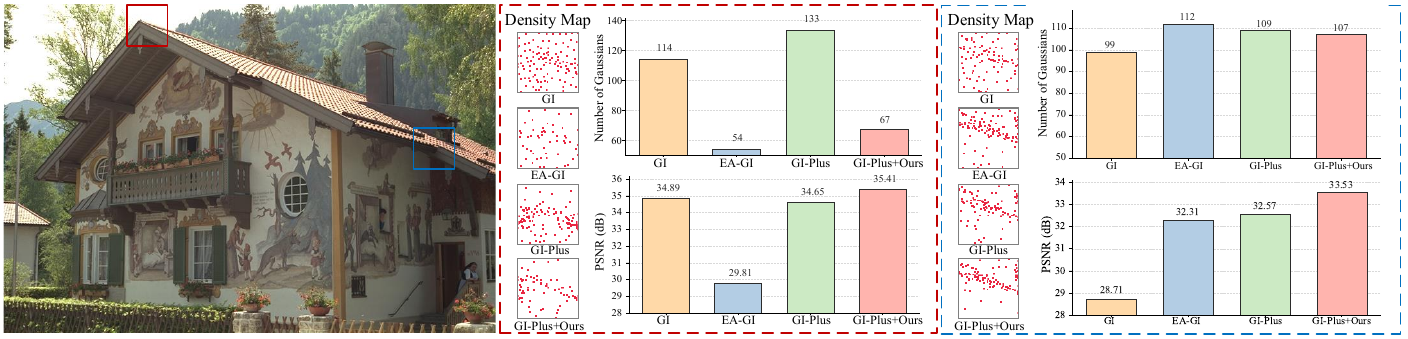}
      \caption{Locality analysis in local regions. For each region, we compare the Gaussian density maps, the number of allocated Gaussians, and the local PSNR. Our method achieves higher local PSNR with fewer or comparable Gaussians, indicating more effective Gaussian allocation according to local image structures.
    }
      \label{Fig: merging_combine}
  \end{figure*}

\subsection{Ablation Study}
\label{sec:ablation-study}
In this section, we conduct ablation experiments to validate the rationality of our proposed method.

\textbf{Effectiveness of each component.}
Tab.~\ref{tab:ablation-psnr} presents the ablation results of RGD and SDGM strategies on three datasets, where SDGM consists of two parts: merging, and color constraint.
The effectiveness of applying the RGD strategy is shown in the second row.
Compared with pixel-wise densification, our approach achieves improvements in PSNR by exploiting the local continuity of image content and reconstruction errors for better structural representation.
The ablation results of SDGM are reported in the third and fourth rows. 
Gaussian merging yields a notable gain in PSNR, indicating that reducing redundant representation in over-reconstructed regions allows more Gaussians to be allocated to complex regions, thereby improving the overall representation capability.
By encouraging better color alignment among neighboring Gaussians before merging, the proposed local color consistency constraint alleviates this issue and further improves reconstruction quality.

\begin{figure}[t]
  \centering
  \includegraphics[width=0.48\textwidth]{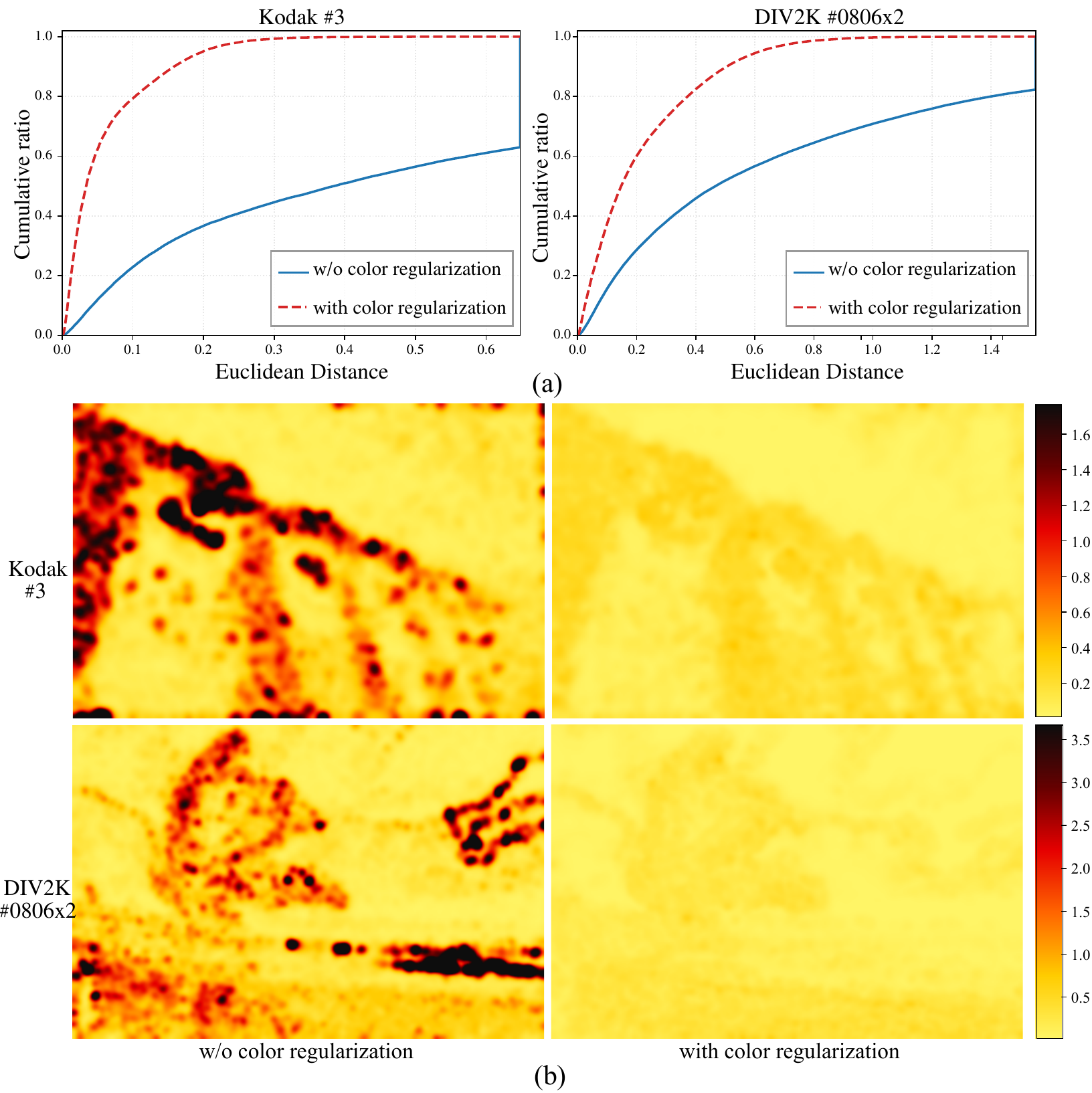}
  \caption{Effectiveness of local color consistency constraint. (a) CDF of pairwise color differences between neighboring Gaussians before merging. (b) Visualization of neighborhood color inconsistency. 
}
  \label{Fig: color_abla}
\end{figure}

\textbf{Locality analysis.}
Fig.~\ref{Fig: merging_combine} presents the Gaussian density maps and quantitative comparisons in two representative local regions. 
A clear observation is that our method achieves higher local reconstruction quality with fewer Gaussians. 
The density maps clearly show that our method allocates Gaussians more efficiently according to local image structure. Specifically, more Gaussians are assigned in structurally complex regions and arranged in a more structure-aligned manner to better capture edges and textures, while in relatively smooth regions, redundant Gaussians are merged to reduce unnecessary representation.
Our LocoADC framework effectively enhances the representation capacity.

\textbf{Effect of local color consistency constraint.}
As shown in Fig.~\ref{Fig: color_abla}(a), applying the color constraint makes the colors of spatially nearby Gaussians more consistent before Gaussian merging. 
The visualization in Fig.~\ref{Fig: color_abla}(b) further confirms that local color fluctuations are effectively reduced. 
This provides a more reliable basis for subsequent Gaussian merging.

\textbf{Comparison with existing Gaussian reduction strategies.}
As discussed in Sec.\ref{sec:density-control-in-3d-gaussian-splatting}, directly adopting sparsification strategies from 3DGS can significantly degrade representation quality. 
To validate this, we compare our method with two Gaussian reduction strategies used in 3DGS for scene reconstruction: size-based pruning\cite{liu2024compgs} and learning-based masking~\cite{fan2024lightgaussian} (Tab.~\ref{tab:ablation existing pruning methods}).
Their inferior performance shows that Gaussian reduction in image representation is fundamentally different from that in scene reconstruction: even small Gaussians may be important for fine-detail representation, and directly removing them would degrade visual quality. 
In contrast, our merging process exploits locality in Gaussian space to identify truly inefficient and neighboring Gaussians, leading to better representation quality.

\begin{table}[t]
  \renewcommand{\arraystretch}{1.2}
  \centering\setlength\tabcolsep{1.0em}
  
  \caption{Comparison with sparsification-based strategies from 3DGS methods on Kodak dataset.}
  \label{tab:ablation existing pruning methods}
  \resizebox{0.8\linewidth}{!}{
  \small 
  \begin{tabular}{ccc}
  \toprule
      Method & PSNR (dB) & MS-SSIM  \\
  \midrule
      Size-based~\cite{liu2024compgs} & 30.5 & 0.9498 \\
      Learning-based~\cite{fan2024lightgaussian} & 32.19 & 0.9527 \\
      SDGM (Ours) & \textbf{32.68} & \textbf{0.9596} \\
  \bottomrule
  \end{tabular}
}
\end{table}

\section{Conclusion}
\label{sec:conclusion}

In this paper, we show that an important source of inefficiency lies in locality-unaware density control: under-reconstructed content is often refined in a fragmented pixel-wise manner, while neighboring optimized Gaussians with similar attributes are redundantly retained. 
To address this issue, we propose LocoADC, a plug-and-play framework that improves Gaussian capacity utilization through Region-wise Gaussian Densification, Similarity-Driven Gaussian merging, and a local color consistency constraint. 
The key idea is to jointly exploit the local continuity of reconstruction errors and the local similarity of neighboring Gaussians for more effective density control.
Experiments on Kodak, DIV2K $\times 2$, and CLIC demonstrate that LocoADC consistently improves GI, EA-GI, and GI-Plus, and comprehensive ablations verify the effectiveness of all components. 
In future work, we plan to extend our method to video representation and 3D scenes, while further formulating density control in a rate-distortion-aware manner.

\section*{Acknowledgements}
This work was supported by the National Natural Science Foundation of China (Grant No.62571160), Engineering Technology R\&D Center of Guangdong Provincial Universities (2024GCZX004), and the Pengcheng Laboratory.
\bibliographystyle{ACM-Reference-Format}
\bibliography{Reference.bib}

\clearpage
\appendix

\begin{figure*}[!t]
  \centering
  \includegraphics[width=1.0\textwidth]{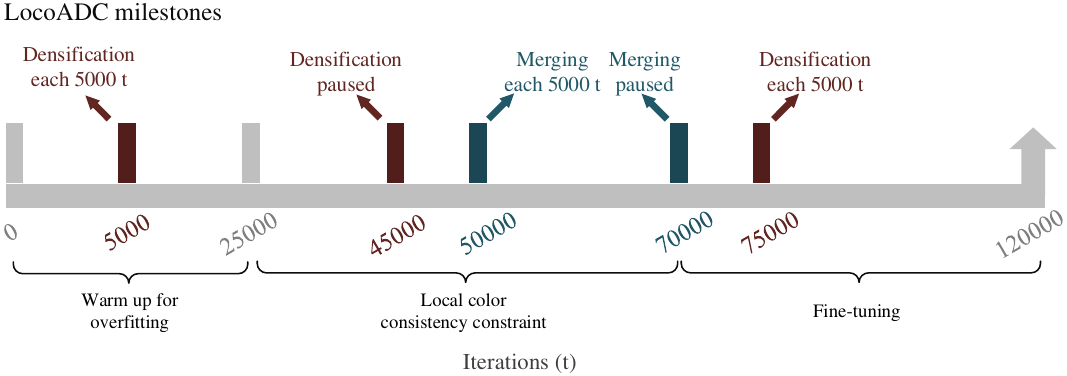}
  \caption{Detailed training process of the proposed LocoADC framework.
}
  \label{Fig: appendix_overview}
\end{figure*}

\section*{Appendix}
We provide more material to supplement our main paper.
This appendix first introduces more implementation details in Sec.~\ref{sec:more-implementation-details}.
Then, we provide additional experimental results in Sec.~\ref{sec:additional-experimental-results}.
Finally, we discuss the limitations and future work in Sec.~\ref{sec:limitations}.

\section{More Implementation Details}
\label{sec:more-implementation-details}

\textbf{Supplementary explanations of LocoADC.}
As shown in Fig.~\ref{Fig: appendix_overview}, the detailed training process of the proposed LocoADC framework is provided.
To avoid introducing improper constraints on Gaussians before they are sufficiently optimized, we enforce the local color consistency constraint only after the warm-up stage.
After the merging operation, the released Gaussian budget is further reallocated through densification to the regions with higher distortion, enabling the model to better fit the image while improving the efficiency of Gaussian representation.

To further facilitate the understanding of our Region-wise Gaussian Densification (RGD) and Similarity-Driven Gaussian Merging (SDGM) strategies, the corresponding pseudocode is provided in Algorithms~\ref{alg:rgd} and~\ref{alg:sdgm}.

\begin{algorithm}[h]
\caption{RGD strategy for $\mathcal{D}(\mathbb{G}_t)$}
\label{alg:rgd}
\begin{algorithmic}[1]
\Require Gaussian set $\mathbb{G}_t$, ground-truth image $I$, rendered image $\hat{I}$, window size $w$, candidate number $K$
\Ensure Updated Gaussian set $\mathcal{D}(\mathbb{G}_t)$

\State $E(x) \gets \|I(x)-\hat{I}(x)\|_2$
\State $\mathbf{u}(x) \gets \dfrac{(\nabla_x E(x), \nabla_y E(x))^\top}{\|\nabla E(x)\|_2^2+\epsilon}$
\State $D(x) \gets \left | \dfrac{1}{w^2}\sum_{q\in N_w(x)} \mathbf{u}(q) \right |_2$
\State $C_K \gets \textsc{TopK}\big(E(x)\cdot D(x)\big)$
\State Group spatially neighboring pixels in $C_K$ into regional subsets
\State Initialize one Gaussian at the center of each region
\State Let all initialized Gaussians form $\Delta \mathbb{G}_t$
\State $\mathbb{G}_t \gets \mathbb{G}_t \cup \Delta \mathbb{G}_t$
\State \Return $\mathbb{G}_t$
\end{algorithmic}
\end{algorithm}


\begin{algorithm}[h]
\caption{SDGM strategy for $\mathcal{S}(\mathbb{G}_t)$}
\label{alg:sdgm}
\begin{algorithmic}[1]
\Require Gaussian set $\mathbb{G}_t$, thresholds $\tau_M$, $\tau_{\text{color}}$, area ratio $\eta$
\Ensure Updated Gaussian set $\mathbb{G}_t$

\State Filter out unstable Gaussians with $\dfrac{g_i}{\max_k g_k} < 0.5$
\State Construct local Gaussian pairs $(i,j)$
\State Retain pairs satisfying $d_M(\mu_i,\mu_j;\Sigma_i) < \tau_M$ and $\|c_i-c_j\|_2 < \tau_{\text{color}}$
\State $D(i,j) \gets \|\mu_i-\mu_j\|_2^2 + \|c_i-c_j\|_2^2 + \|\ell_i-\ell_j\|_2^2$
\State For each Gaussian involved in multiple pairs, retain the pair with the lowest $D(i,j)$
\For{each selected pair $(i,j)$}
    \State Compute the merged Gaussian $g_{\text{merged}}$
    \If{$S_{\text{merged}} \le \eta(S_i+S_j)$}
        \State Replace $g_i$ and $g_j$ in $\mathbb{G}_t$ with $g_{\text{merged}}$
    \EndIf
\EndFor
\State Let the updated Gaussian set form $\mathbb{G}_t$
\State \Return $\mathbb{G}_t$
\end{algorithmic}
\end{algorithm}

\textbf{Effective support area.}
In Eq.~\ref{eq:merging weight}, we impose a support-area constraint based on the effective support areas $S_i$ of the original and merged Gaussians. 
We provide the definition of the effective support area used in that formulation.
The influence of a 2D Gaussian is primarily concentrated in a local neighborhood around its mean. 
To quantify this practical spatial extent, for a Gaussian \(g_i\) with mean 
\(\mu_i \in \mathbb{R}^2\) and covariance \(\Sigma_i \in \mathbb{R}^{2\times2}\), we define 
its effective support region as
\begin{equation}
   \mathcal{A}_i = \left\{ x \in \mathbb{R}^2 \;\middle|\;
(x-\mu_i)^\top \Sigma_i^{-1}(x-\mu_i) \le \tau
\right\} 
\end{equation}
where \(\tau > 0\) is a predefined Mahalanobis-distance threshold. The corresponding 
effective support area is defined as
\begin{equation}
    S_i = \mathrm{Area}(\mathcal{A}_i)
\end{equation}
Since \(\mathcal{A}_i\) is an equal-density ellipse induced by \(\Sigma_i\), its area can be written as:
\begin{equation}
    S_i = \pi \tau \sqrt{|\Sigma_i|}
\end{equation}
where \(|\Sigma_i|\) denotes the determinant of \(\Sigma_i\). Therefore, \(S_i\) provides a compact quantitative measure of the practical spatial coverage of Gaussian \(g_i\), which is used in Eq.~\ref{eq:merging weight} to compute the Gaussian merging weights.
Note that, in the actual computation of Eq.~\ref{eq:merging weight}, the constant factor \(\pi\tau\) is canceled out. 
Therefore, in implementation, we only use the proportional term \(\sqrt{|\Sigma_i|}\) 
without explicitly evaluating the constant.
  
\textbf{Hyperparameter setting.}
Unless otherwise specified, we use the following default hyperparameters for LocoADC.
In RGD strategy, the default  window size is $w=7$ (i.e., a $7\times7$ local window) in Eq.~\ref{eq:local coherence}.
For similar Gaussian pair selection, the Mahalanobis overlap threshold is set to $\tau_{M}=5.0$, and the color similarity threshold is set to $\tau_{\mathrm{color}} = 10/255$ (RGB $\ell_2$ distance).
The merging constraint ratio is $\eta=1.5$.
For local color consistency constraint, we enable color constraint with weight
$\lambda_{\mathrm{color}}=0.01$.

\begin{figure*}[t]
  \centering
  \includegraphics[width=0.95\textwidth]{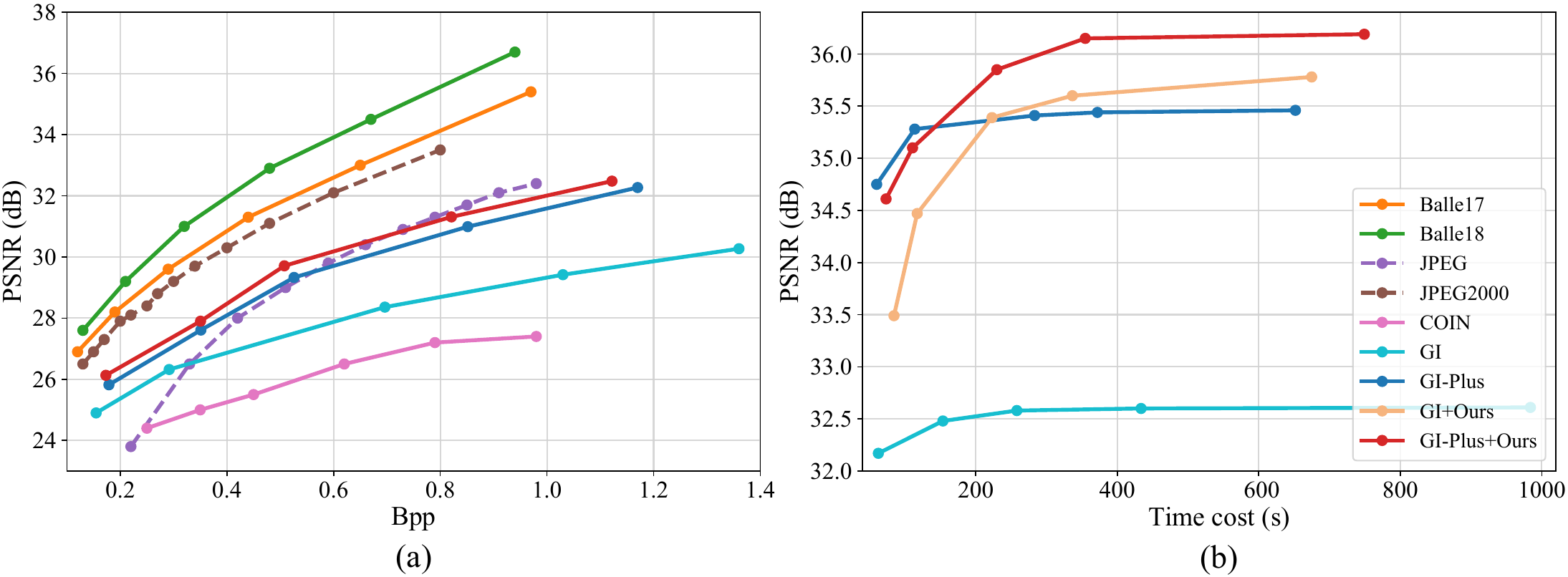}
  \caption{Rate-distortion (left) and Time-PSNR (right) comparisons.
}
  \label{Fig: color_abla}
\end{figure*}

\section{Additional Experimental Results}
\label{sec:additional-experimental-results}

\textbf{Rate-Distortion and Time-Quality Analysis.}
To further evaluate the representation efficiency of our method, we compare it with traditional image codecs~\cite{wallace1991jpeg, taubman2002jpeg2000}, learned image compression methods~\cite{balle2016end, balle2018variational}, an INR-based method~\cite{dupont2021coin}, and representative Gaussian-based image representation methods~\cite{zhang2024gaussianimage, li2026gaussianimage++}.
The PSNR-Bpp comparison in Fig.~\ref{Fig: color_abla}(a) shows that our method consistently achieves a better rate-distortion trade-off than existing Gaussian-based approaches while approaching the performance of existing learned image compression methods.
Fig.~\ref{Fig: color_abla}(b) further reports the PSNR-Time curves.
Although our method converges slightly slower in the early stage due to the additional optimization introduced by RGD and SDGM, it consistently surpasses the baseline after convergence, demonstrating a better quality-efficiency trade-off.

\begin{table}[t] 
\centering 
\caption{Recent GS-based method comparisons on Kodak. GS Num. denotes the number of Gaussians.} 
\label{tab:additional_sota_comparison} 
\renewcommand{\arraystretch}{1.2} 
\setlength{\tabcolsep}{5pt} 
\resizebox{0.95\linewidth}{!}{ 
\begin{tabular}{l|cccc} 
\toprule 
Methods & PSNR $\uparrow$ & MS-SSIM $\uparrow$ & Params $\downarrow$ & GS Num. \\
\midrule 
SmartSplat~\cite{li2026smartsplat} & 33.47 & 0.9813 & 0.08M & 10k \\ 
LIG~\cite{zhu2025large} & 31.00 & 0.9750 & 0.08M & 10k \\ 
Fast2DGS~\cite{wang2026fast} & 33.79 & -- & 0.08M & 10k \\ 
GI-Plus+Ours & {36.15} & 0.9826 & 0.08M & 10k \\ 
\cmidrule(lr){1-5} 
StructureGI~\cite{liang2025structure} & 45.40 & 0.9987 & 0.56M & 70k \\ 
GI-Plus+Ours & {46.83} & {0.9991} & {0.38M} & 48k \\ 
\bottomrule 
\end{tabular} 
} 
\end{table}

\textbf{Comparison with Recent GS-based Image Representation Methods.}
Tab.~\ref{tab:additional_sota_comparison} reports comparisons with recent Gaussian-based image representation methods~\cite{li2026smartsplat, zhu2025large, wang2026fast, liang2025structure}.
Our method consistently achieves the best performance under the same Gaussian budget.
Compared with StructureGI~\cite{liang2025structure}, our method further achieves superior reconstruction quality with significantly fewer Gaussians and model parameters.

\textbf{Effectiveness of RGD strategy.}
To verify the effectiveness of the proposed RGD strategy, we further compare it with a magnitude-only regional densification variant. 
Specifically, this variant also performs region-wise Gaussian allocation, but identifies candidate regions solely according to the distortion magnitude, without considering the local distortion coherence defined in Eq.~(\ref{eq:local coherence}).

\begin{table}[h]
  \centering
  \small
  \setlength{\tabcolsep}{6pt}
  \renewcommand{\arraystretch}{1.2}
  \caption{Ablation study on RGD strategy on Kodak dataset.}
  \begin{tabular}{ccc}
    \toprule
    Variant & PSNR (dB) & MS-SSIM \\
    \midrule
    Pixel-wise Densification & 31.9 & 0.9614 \\
    Magnitude-only Region-wise Densification & 31.98 & 0.9617 \\
    RGD (Ours) & \textbf{32.24} & \textbf{0.964} \\
    \bottomrule
  \end{tabular}
  \label{tab:ablation-rgd-variant}
\end{table}

\begin{figure}[t]
  \centering
  \includegraphics[width=0.48\textwidth]{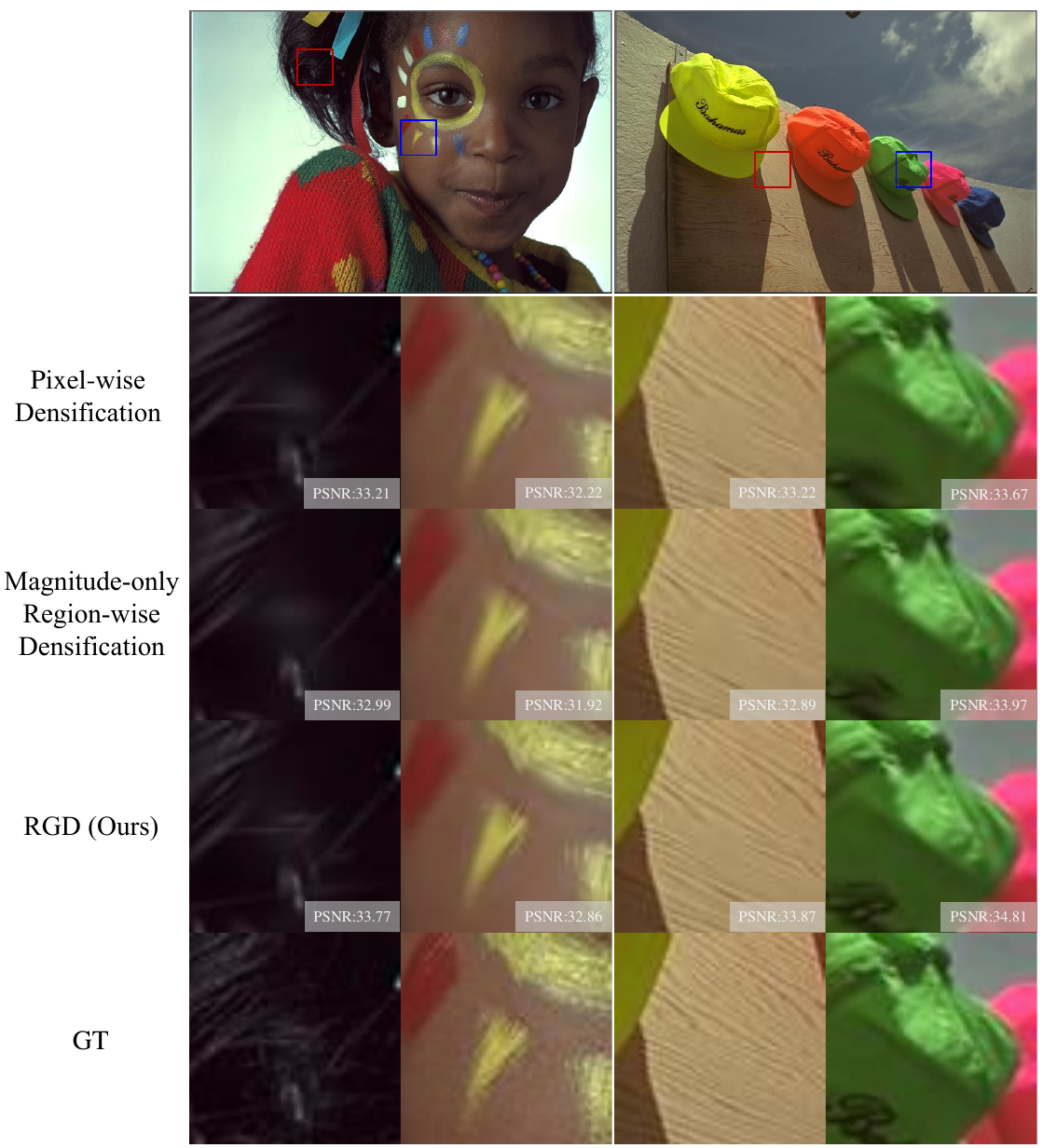}
  \caption{Qualitative ablation of the RGD strategy. Compared with pixel-wise densification and magnitude-only region-wise densification, our method better recovers coherent fine structures and textured details. 
}
  \label{fig:growing}
\end{figure}

As reported in Tab.~\ref{tab:ablation-rgd-variant}, the magnitude-only variant brings only marginal improvement over pixel-wise densification, increasing PSNR from 31.90 to 31.98 and MS-SSIM from 0.9614 to 0.9617. 
In contrast, our full RGD strategy achieves the best performance. 
This result indicates that the effectiveness of RGD arises not only from region-wise densification itself, but also from the proposed coherence-aware region identification.
The qualitative comparisons in Fig.~\ref{fig:growing} further support this observation, as our method recovers finer structures and more faithful textures than the magnitude-only variant.

\textbf{Hyper-parameter sensitivity analysis.}
To investigate the robustness of LocoADC to different hyper-parameters, we conduct a one-factor-at-a-time sensitivity analysis on Kodak.
Specifically, we vary one hyper-parameter while keeping all others fixed to the default setting.
As reported in Tab.~\ref{tab:hyper_sensitivity}, the performance remains stable within reasonable ranges of the tested parameters, indicating that LocoADC does not rely on carefully tuned hyper-parameters.
We also observe that overly loose merging-related thresholds, such as large $\tau_{\text{color}}$ and $\eta$, may slightly reduce MS-SSIM, since they allow more aggressive merging and may introduce mild smoothing.
Therefore, we adopt a moderate global setting for all images and datasets in our default implementation.


\begin{table}[h] 
\centering \caption{Hyper-parameter sensitivity analysis on Kodak.} 
\label{tab:hyper_sensitivity} 
\renewcommand{\arraystretch}{1.2} 
\setlength{\tabcolsep}{4.5pt} 
\resizebox{0.95\linewidth}{!}{ 
\begin{tabular}{c|c|ccccc} \toprule Hyper-parameter & Metric & \multicolumn{5}{c}{Values} \\ 
\midrule 
\multirow{3}{*}{$\lambda_{\text{color}}$} & Value & 0.001 & 0.005 & 0.01 & 0.02 & 0.05 \\ 
& PSNR $\uparrow$ & 32.59 & 32.64 & 32.68 & 32.62 & 32.61 \\ 
& MS-SSIM $\uparrow$ & 0.9588 & 0.9591 & 0.9596 & 0.9590 & 0.9590 \\ 
\midrule 
\multirow{3}{*}{$w$} & Value & 3 & 5 & 7 & 9 & 11 \\ 
& PSNR $\uparrow$ & 32.67 & 32.66 & 32.68 & 32.65 & 32.64 \\ 
& MS-SSIM $\uparrow$ & 0.9594 & 0.9593 & 0.9596 & 0.9593 & 0.9593 \\ 
\midrule \multirow{3}{*}{Gradient cutoff} & Value & 0.1 & 0.3 & 0.5 & 0.7 & 0.9 \\ 
& PSNR $\uparrow$ & 32.59 & 32.66 & 32.68 & 32.63 & 32.55 \\ 
& MS-SSIM $\uparrow$ & 0.9622 & 0.9603 & 0.9596 & 0.9590 & 0.9582 \\ 
\midrule 
\multirow{3}{*}{$\tau_M$} & Value & 1 & 3 & 5 & 7 & 9 \\ 
& PSNR $\uparrow$ & 32.53 & 32.62 & 32.68 & 32.64 & 32.59 \\ 
& MS-SSIM $\uparrow$ & 0.9628 & 0.9614 & 0.9596 & 0.9592 & 0.9587 \\ 
\midrule 
\multirow{3}{*}{$\tau_{\text{color}}$} & Value & 2.5 & 5 & 10 & 15 & 20 \\ 
& PSNR $\uparrow$ & 32.63 & 32.65 & 32.68 & 32.71 & 32.72 \\ 
& MS-SSIM $\uparrow$ & 0.9624 & 0.9611 & 0.9596 & 0.9584 & 0.9579 \\ 
\midrule 
\multirow{3}{*}{$\eta$} & Value & 1 & 1.5 & 3 & 5 & -- \\ 
& PSNR $\uparrow$ & 32.67 & 32.68 & 32.61 & 32.47 & -- \\ 
& MS-SSIM $\uparrow$ & 0.9595 & 0.9596 & 0.9578 & 0.9556 & -- \\ 
\bottomrule 
\end{tabular} 
} 
\end{table}

\textbf{Effectiveness of each component.}
Tab.~\ref{tab:ablation-time} reports the reconstruction quality and training time of different module combinations.
RGD improves both PSNR and MS-SSIM by allocating additional Gaussians to under-reconstructed regions, thereby correcting local reconstruction errors and enhancing structural consistency.
In contrast, merging further improves PSNR by removing redundant Gaussians, while slightly reducing MS-SSIM because consolidating similar Gaussians in locally smooth regions may mildly smooth subtle edge variations.
The local color consistency constraint alleviates this effect by encouraging neighboring Gaussians to maintain consistent colors, leading to further improvements in both PSNR and MS-SSIM.
Overall, the proposed methods achieve the best reconstruction quality with only a moderate increase in training time.

\textbf{Comparison with 3DGS density control strategies.}
Tab.~\ref{tab:3dgs_densification_comparison} compares our method with representative 3DGS density control strategies~\cite{kerbl20233d, ye2024absgs, hu2024gauhuman}.
These methods are less effective for image representation because position gradients are typically too weak to reliably identify under-reconstructed regions for densification.
Consequently, our distortion-guided density control achieves substantially better reconstruction quality.


\begin{table}[t]
  \centering
  \small
  \setlength{\tabcolsep}{4.5pt}
  \renewcommand{\arraystretch}{1.2}
  \caption{Training time analysis on RGD (Sec.~\ref{sec:structure-guided-regional-densification}) and SDGM (Sec.~\ref{sec:similarity-driven-gaussian-consolidation}) strategies. The GI-Plus is selected as the baseline.}
  \label{tab:ablation-time}
  \resizebox{.48\textwidth}{!}{%
  \begin{tabular}{cccc|ccc}
    \toprule
    GI-Plus & RGD & Merging & Color Cons. & PSNR $\uparrow$ & MS-SSIM $\uparrow$ & Time (s) \\
    \midrule
    $\checkmark$ &  &  &  & 31.90 & 0.9614 & 247.75 \\
    $\checkmark$ & $\checkmark$ &  &  & 32.24 & 0.9640 & 260.55 \\
    $\checkmark$ &  & $\checkmark$ &  & 32.34 & 0.9561 & 268.2 \\
    $\checkmark$ &  & $\checkmark$ & $\checkmark$ & 32.52 & 0.9577 & 302.47 \\
    \midrule
    $\checkmark$ & $\checkmark$ & $\checkmark$ & $\checkmark$ & {32.68} & 0.9596 & 313.88 \\
    \bottomrule
  \end{tabular}
  }
\end{table}

\begin{table}[t]
\centering
\caption{3DGS density control comparison.}
\label{tab:3dgs_densification_comparison}
\renewcommand{\arraystretch}{1.0}
\setlength{\tabcolsep}{4pt}
\resizebox{0.6\linewidth}{!}{
\begin{tabular}{l|cc}
\toprule
Method & PSNR $\uparrow$ & MS-SSIM $\uparrow$ \\
\midrule
3DGS~\cite{kerbl20233d} & 28.99 & 0.9435 \\
AbsGS~\cite{ye2024absgs} & 28.93 & 0.9430 \\
GauHuman~\cite{hu2024gauhuman} & 28.85 & 0.9426 \\
Ours & 32.68 & 0.9596 \\
\bottomrule
\end{tabular}
}
\end{table}

\begin{figure*}[t]
  \centering
  \includegraphics[width=0.85\textwidth]{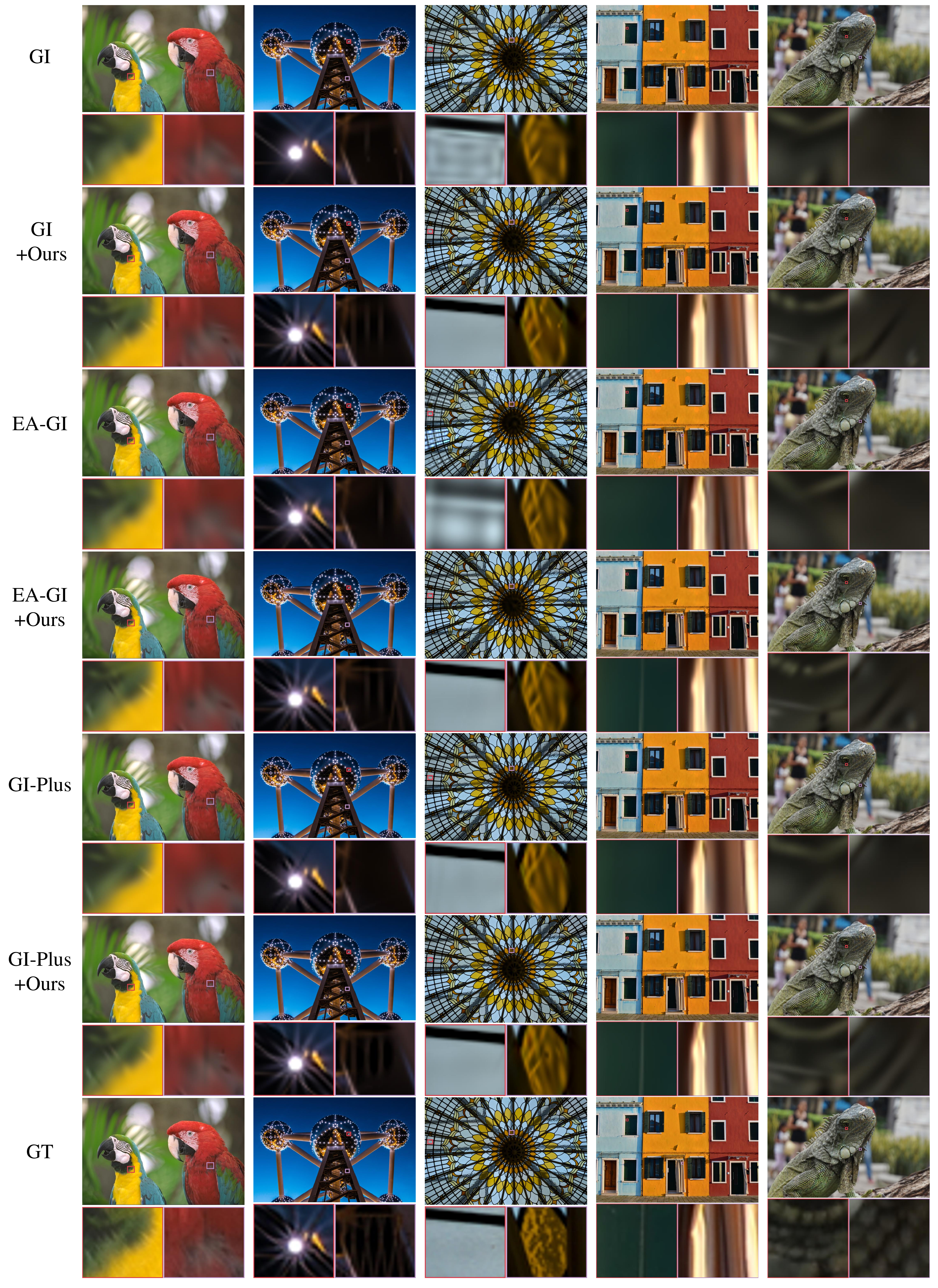}
  \caption{Qualitative results on Kodak, DIV2k $\times 2$ and CLIC datasets.
}
  \label{Fig: appendix_comparison}
\end{figure*}

\textbf{More Qualitative Results.}
Fig.~\ref{Fig: appendix_comparison} provides more qualitative comparisons among GI~\cite{zhang2024gaussianimage}, EA-GI~\cite{chen2026entropy}, and GI-Plus~\cite{li2026gaussianimage++}, as well as their counterparts equipped with our LocoADC framework.
The results demonstrate that applying our LocoADC framework consistently improves the visual quality over the baselines.

\section{Limitations}
\label{sec:limitations}
Despite the promising results achieved by our method for Gaussian image representation, several limitations remain. 
First, there is still a noticeable performance gap between Gaussian-based methods and neural image representations.
Second, the current framework is mainly validated on conventional image settings, while its applicability to more challenging formats, such as \(360^\circ\) and panoramic images, remains underexplored. 
Finally, this work focuses on image representation and reconstruction, and the potential of Gaussian image representations for downstream tasks, such as image classification, object detection, and style transfer, has yet to be investigated.

\end{document}